\documentclass[sn-mathphys-num]{sn-jnl}

\usepackage{graphicx}%
\usepackage{multirow}%
\usepackage{amsmath,amssymb,amsfonts}%
\usepackage{amsthm}%
\usepackage{mathrsfs}%
\usepackage[title]{appendix}%
\usepackage{xcolor}%
\usepackage{textcomp}%
\usepackage{manyfoot}%
\usepackage{booktabs}%
\usepackage{algorithm}%
\usepackage{algorithmicx}%
\usepackage{algpseudocode}%
\usepackage{listings}%

\usepackage{multicol}
\usepackage{makecell}
\usepackage{colortbl}
\usepackage{caption}

\begin{document}

\title{Distribution Aligned Semantics Adaption for Lifelong Person Re-Identification}


\author[1]{\fnm{Qizao} \sur{Wang}}\email{qzwang22@m.fudan.edu.cn}

\author[2]{\fnm{Xuelin} \sur{Qian}}\email{xlqian@nwpu.edu.cn}

\author[1]{\fnm{Bin} \sur{Li}}\email{libin@fudan.edu.cn}

\author*[1]{\fnm{Xiangyang} \sur{Xue}}\email{xyxue@fudan.edu.cn}

\affil[1]{\orgname{Fudan University}, \orgaddress{\city{Shanghai}, \country{China}}}

\affil[2]{\orgname{Northwestern Polytechnical University}, \orgaddress{\city{Xi'an}, \country{China}}}

\abstract{
In real-world scenarios, person Re-IDentification (Re-ID) systems need to be adaptable to changes in space and time. Therefore, the adaptation of Re-ID models to new domains while preserving previously acquired knowledge is crucial, known as Lifelong person Re-IDentification (LReID). Advanced LReID methods rely on replaying exemplars from old domains and applying knowledge distillation in logits with old models. However, due to privacy concerns, retaining previous data is inappropriate. Additionally, the fine-grained and open-set characteristics of Re-ID limit the effectiveness of the distillation paradigm for accumulating knowledge. We argue that a Re-ID model trained on diverse and challenging pedestrian images at a large scale can acquire robust and general human semantic knowledge. These semantics can be readily utilized as shared knowledge for lifelong applications. In this paper, we identify the challenges and discrepancies associated with adapting a pre-trained model to each application domain and introduce the Distribution Aligned Semantics Adaption (DASA) framework. It efficiently adjusts Batch Normalization (BN) to mitigate interference from data distribution discrepancy and freezes the pre-trained convolutional layers to preserve shared knowledge. Additionally, we propose the lightweight Semantics Adaption (SA) module, which effectively adapts learned semantics to enhance pedestrian representations. Extensive experiments demonstrate the remarkable superiority of our proposed framework over advanced LReID methods, and it exhibits significantly reduced storage consumption. DASA presents a novel and cost-effective perspective on effectively adapting pre-trained models for LReID. The code is available at \url{https://github.com/QizaoWang/DASA-LReID}.
}


\maketitle

\section{Introduction}
\label{sec:intro}
Person Re-IDentification (Re-ID) aims at recognizing the same pedestrian across disjoint cameras. With significant advancements in deep learning algorithms over the past decade, it has demonstrated remarkable performance~\cite{wang2018learning,qian2019leader,zheng2019joint,wang2022co,wang2023rethinking,wang2024exploring,wang2024large}. 
However, advanced methods usually assume the pedestrian data is provided at once. In real-world scenarios, surveillance data accumulates continuously due to successive deployments and ongoing applications. Re-ID data expands discretely, both spatially and temporally, necessitating the adaptation of Re-ID models to new environments. Consequently, Lifelong person Re-IDentification (LReID) seeks to continuously adapt Re-ID models to novel domains while preserving previously acquired knowledge during the incremental learning process.
Unfortunately, due to substantial variations between different Re-ID domains, merely updating the model with new surveillance data can lead to a loss of discriminative ability in previously learned domains. This phenomenon is commonly referred to as the catastrophic forgetting problem.

\begin{figure}[t]
\centering
  \includegraphics[width=\linewidth]{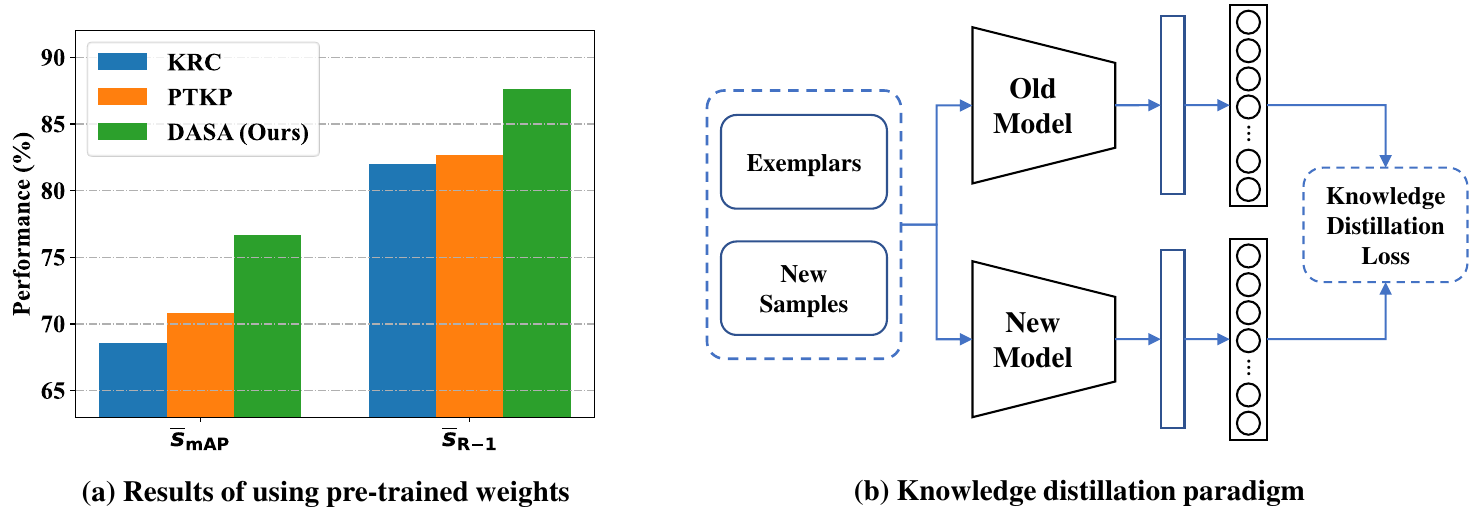}
  \caption{(a) Comparison of average incremental performance for different methods using the same pre-trained weights. (b) Conventional knowledge distillation pipeline with exemplars. Our proposed DASA paradigm shows great superiority in LReID.}
  \label{fig:intro}
\end{figure}

Recently, we have witnessed the remarkable success of pre-training in computer vision~\cite{he2022masked,Kirillov_2023_ICCV}, which also revolutionizes the field of person Re-ID. For example, \citet{fu2021unsupervised,fu2022large}~adopt unsupervised pre-training to prevailing Convolutional Neural Network (CNN) architectures like ResNet~\cite{resnet} and achieve significant performance improvement. It is promising to leverage pre-trained models that can be obtained effortlessly for real-world applications.
Nevertheless, as observed in Fig.~\ref{fig:intro}~(a), directly using the pre-trained weights for model initialization yields limited benefits in the lifelong evolution scenario. Advanced LReID methods~\cite{pu2021lifelong,ge2022lifelong,yu2023lifelong,wu2021generalising} follow the prevailing knowledge distillation paradigm~\cite{li2017learning}. Thus we are motivated to think whether the performance degradation stems from the inherent drawbacks of the knowledge distillation paradigm.

As depicted in Fig.~\ref{fig:intro}~(b), existing LReID methods follow the distillation baseline to distill knowledge at the logit level. This paradigm heavily relies on the learned classifier.
However, unlike conventional lifelong learning~\cite{kirkpatrick2017overcoming,li2017learning,tung2019similarity,zhao2021continual}, LReID is inherently a fine-grained open-set problem. The distillation paradigm would restrict the model's ability to incorporate valuable knowledge for recognizing unseen identities during inference. It also limits person Re-ID in real-world applications in two aspects. 
Firstly, relying on replaying data of old tasks, namely exemplars~\cite{wu2019large,zhao2020maintaining,rebuffi2017icarl,hou2019learning}, is impractical due to strict privacy constraints associated with pedestrian data.
Secondly, the number of classes in the LReID datasets significantly exceeds those in conventional lifelong learning tasks, such as ImageNet~\cite{deng2009imagenet}. The high dimensional classifier would consume great resources when saving it for knowledge distillation.

Therefore, to achieve a balance between knowledge preservation and updating with minimal source consumption, and without using exemplars, we introduce the Distribution Aligned Semantics Adaption (DASA) framework, which presents a novel LReID paradigm based on pre-trained models. 
In this paradigm, we do not rely on knowledge distillation that uses saved old data and models for knowledge retention. Instead, we exert the advantages of the pre-trained model in the lifelong learning process. Specifically, we carefully handle and leverage the two crucial techniques in CNN, \textit{i.e.}, Batch Normalization (BN) and Convolutional (Conv) blocks.
Firstly, BN effectively regularizes the model and captures data distributions, contributing to improved learning~\cite{zhuang2020rethinking,seo2020learning}. In our proposed paradigm, we tune BN effectively to deal with domain distribution differences between pre-training and application.
Secondly, Conv layers specialize in encoding data patterns and semantics at different depths of the network. It is expected that the Conv layers trained on a substantial amount of pedestrian data can acquire robust and generalizable human semantic knowledge~\cite{fu2021unsupervised,fu2022large}, thereby serving as a valuable source of shared knowledge across domains. Therefore, we propose freezing the pre-trained Conv layers to preserve shared person semantic knowledge. However, general semantics are not capable of distinguishing pedestrians in complex application scenarios. To adapt the acquired general knowledge to each application domain effectively, we introduce the lightweight Semantics Adaption (SA) module, which can efficiently aggregate, refine, and generate improved pedestrian representations. Adapting the general human semantics to more representative identity semantics represents a process from coarse to fine, showing great ability to evolve the model for lifelong application.
\textbf{Our contributions can be summarized as follows:}

\textbf{(1)} We advocate taking advantage of the robust and general human semantic knowledge acquired through large-scale pre-training to effectively adapt models in application domains, presenting a new paradigm free of exemplars for LReID.

\textbf{(2)} We propose the Distribution Aligned Semantics Adaption (DASA) framework for LReID. It eliminates the interference of domain distribution discrepancy between pre-training and application, and incorporates the lightweight semantics adaption module to aggregate and refine learned semantics for better pedestrian representations.

\textbf{(3)} Extensive experiments demonstrate the superiority of the proposed framework for LReID, achieving state-of-the-art results with significantly reduced resource consumption compared to other advanced LReID methods.

\section{Related Work}
\label{sec:related_work}
\subsection{Lifelong Learning}
Lifelong learning, also known as incremental or continual learning, seeks to maintain stable performance on old tasks while adapting the model to gain new knowledge. Methods can be traditionally divided into three categories, namely regularization-based, rehearsal-based, and architecture-based methods.
Regularization-based methods~\cite{kirkpatrick2017overcoming,li2017learning} limit updating important parameters for preceding tasks to mitigate forgetting.
To better keep past acquired knowledge, rehearsal-based methods~\cite{rebuffi2017icarl,wu2019large,hou2019learning} maintain a memory buffer to store finite exemplars of previous tasks. They are prevalent for their satisfactory performance and simplicity.
However, rehearsal-based methods generally deteriorate with a smaller buffer size and do not apply to scenarios where data privacy should be considered carefully.
Architecture-based methods design separate components or extra parameters for new tasks, so they are immune to forgetting. They expand the network with task-specific components~\cite{li2019learn,yoon2018lifelong} or attend to task-specific sub-networks~\cite{mallya2018packnet,serra2018overcoming}.
For instance, DER~\cite{yan2021dynamically} dynamically adds a new feature extractor per task, and HAT~\cite{serra2018overcoming} learns a hard attention mask concurrently to every task. 
Although lifelong learning has received extensive research in classification or recognition tasks, the presence of imbalanced samples per identity and subtle inter-class variations in person Re-ID poses additional challenges. The significant variations across domains make the problem of catastrophic forgetting even more thorny.

\subsection{Lifelong Person Re-Identification}
Recently, there has been remarkable progress in person re-identification leveraging pre-prepared stationary training data~\cite{wang2018learning,luo2019bag,zheng2019joint,wang2023rethinking,meng2024unleashing,wang2024content}. In response to the demand for long-term scenarios, \citet{pu2021lifelong} introduce Lifelong person Re-IDentification (LReID) and propose to maintain a learnable knowledge graph to adaptively update previous knowledge. However, due to significant domain variations, it struggles to retain old knowledge without access to previous data.
Recently, state-of-the-art LReID methods have drawn inspiration from rehearsal-based lifelong learning approaches. They adopt knowledge distillation baselines designed for traditional lifelong learning to preserve acquired knowledge. 
For instance, based on the distillation paradigm, GwFReID~\cite{wu2021generalising} formulates a comprehensive learning objective for maintaining coherence during progressive learning. PTKP~\cite{ge2022lifelong} proposes a pseudo task knowledge preservation framework to alleviate the domain gap in the last BN layer. KRC~\cite{yu2023lifelong} introduces a dynamic memory model for bi-directional knowledge transfers and a knowledge consolidation scheme.
However, the intrinsic dissimilarities between Re-ID with the classification task restrict these methods from effectively incorporating and leveraging useful knowledge for LReID. Moreover, privacy concerns limit their applicability in real-world scenarios, where exemplars cannot be stored for lifelong usage. The recent work Teata~\cite{wang2024image} achieves knowledge alignment, transfer, and accumulation within an ``image-text-image'' closed loop, leveraging the advantages of text semantics while avoiding reliance on exemplars.

In LReID, we aim to achieve a balance between knowledge preservation and updating with minimal source consumption, and without using exemplars. It differs from existing studies in Test-Time Adaption (TTA)~\cite{wangtent,han2022generalizable} and Domain Generation (DG)~\cite{jin2020style,liu2022debiased}. 
Specifically, in TTA, the pre-trained model is optimized during inference \cite{wangtent}. For instance, BNTA~\cite{han2022generalizable} adapts the model using gallery data and self-supervised auxiliary tasks, and TEMP~\cite{adachi2024test} uses query images and gallery features to minimize entropy at test time. Differently, in LReID, the model is updated during training and directly tested on seen and unseen domains. 
In DG, studies aim to improve the generalization performance of the model on unseen domains~\cite{jin2020style,liu2022debiased}, and the model is usually trained on multiple domains to learn domain-invariant representations. In contrast, in LReID, the model is optimized with various domains sequentially and the data from previous domains is inaccessible, so knowledge forgetting will be a great challenge.

\section{Methodology}
\label{sec:method}

\subsection{Preliminary}
\noindent \textbf{Problem formulation.}
In LReID, a stream of datasets $D = {\{ D^{(t)} \}}^{T}_{t=1}$ are used for model training sequentially. Each dataset is regarded as an application domain during the lifelong deployment of the Re-ID model. At the $t$-th step, $D^{(t)}=\{ D^{(t)}_{train}, D^{(t)}_{test} \}$ contains the training set and the testing set.
$D^{(t)}_{train} = {\{ (x^{(t)}_{i}, y^{(t)}_{i})  \}}^{n^{(t)}}_{i=1}$, where $x^{(t)}_{i}$ and $y^{(t)}_{i} \in [1, n^{(t)}_{id}]$ are the $i$-th person image and its identity label, respectively, where $n^{(t)}_{id}$ is the total number of identity classes in the $t$-th training set, and $n^{(t)}$ is the total number of samples in the $t$-th training set. 
At the $t$-th training step, the model $\mathcal{G}^{(t)}$ as feature extractor and a classifier $g^{(t)}$ are updated with $D^{(t)}_{train}$. 
Since the identity classes for training and testing are disjoint, the classifier is discarded and the trained model is used for evaluation. 
The model $\mathcal{G}^{(t)}$ is expected to perform well on all testing sets of seen domains, \textit{i.e.}, ${\{ D^{(t')}_{test} \}}^{t}_{t'=1}$, respectively. The extracted feature of an input image $x_{i}$ using $\mathcal{G}^{(t)}$ is denoted as $f^{(t)}_{i} \in \mathbb{R}^{d}$, where $d$ denotes the feature dimension. For notation simplicity, we omit the subscript $i$ in the following.

\noindent \textbf{Knowledge distillation paradigm.}
\label{subsec:knowledge_distillation} 
The serious challenge of diverse variation across different datasets obliges previous LReID methods~\cite{wu2021generalising,yu2023lifelong,ge2022lifelong} to save a small number of exemplars from previous training steps and use them for knowledge distillation~\cite{li2017learning,rebuffi2017icarl}.
In the knowledge distillation paradigm, $\mathcal{G}^{(t-1)}$ and $g^{(t-1)}$ are used to maintain the acquired knowledge. The distillation is performed between the old $\mathcal{G}^{(t-1)}$, $g^{(t-1)}$ and the evolving $\mathcal{G}^{(t)}$, $g^{(t)}$.
We denote the extracted feature of an exemplar image $x^{r}$ using $\mathcal{G}^{(t)}$ as ${f^{r}}^{(t)} \in \mathbb{R}^{d}$, and the weight parameter of the classifier $g^{(t)}$ as ${\phi}^{(t)} \in \mathbb{R}^{N^{(t)} \times d}$, where $N^{(t)} = \sum_{j=1}^{t} n^{(j)}_{id}$. The knowledge distillation loss is formulated as follows:
\begin{equation}
\mathcal{L}_{kd} = - \frac{1}{|\mathcal{B}|} \sum\limits_{x^{r} \in \mathcal{B}} \sigma \big({f^{r}}^{(t-1)} \cdot {{\phi}^{(t-1)}}^{\top} \big) \ \log \left( \sigma \big({f^{r}}^{(t)} \cdot {{\phi}^{(t)}}^{\top} \big) \right),
\label{eq:L_kd}
\end{equation}
\noindent where $|\mathcal{B}|$ denotes the training batch size, $\sigma$ is the $softmax$ function. For dimension consistency of ${\phi}^{(t-1)}$ and ${\phi}^{(t)}$ in Eq.~\ref{eq:L_kd}, either only old classes of $g^{(t)}$ are used, or $g^{(t-1)}$ is expanded by using $\mathcal{G}^{(t)}$ to calculate the center of the new classes before the new training step. New samples can also be used for distillation~\cite{li2017learning,pu2021lifelong}.
Previous LReID works~\cite{ge2022lifelong,yu2023lifelong,wu2021generalising} all follow the knowledge distillation paradigm, but the significant task discrepancy between classification and Re-ID makes them suboptimal.
Due to privacy issues, the training data from previous steps should not be available anymore.

\begin{figure*}[t]
  \centering
  \includegraphics[width=0.98\linewidth]{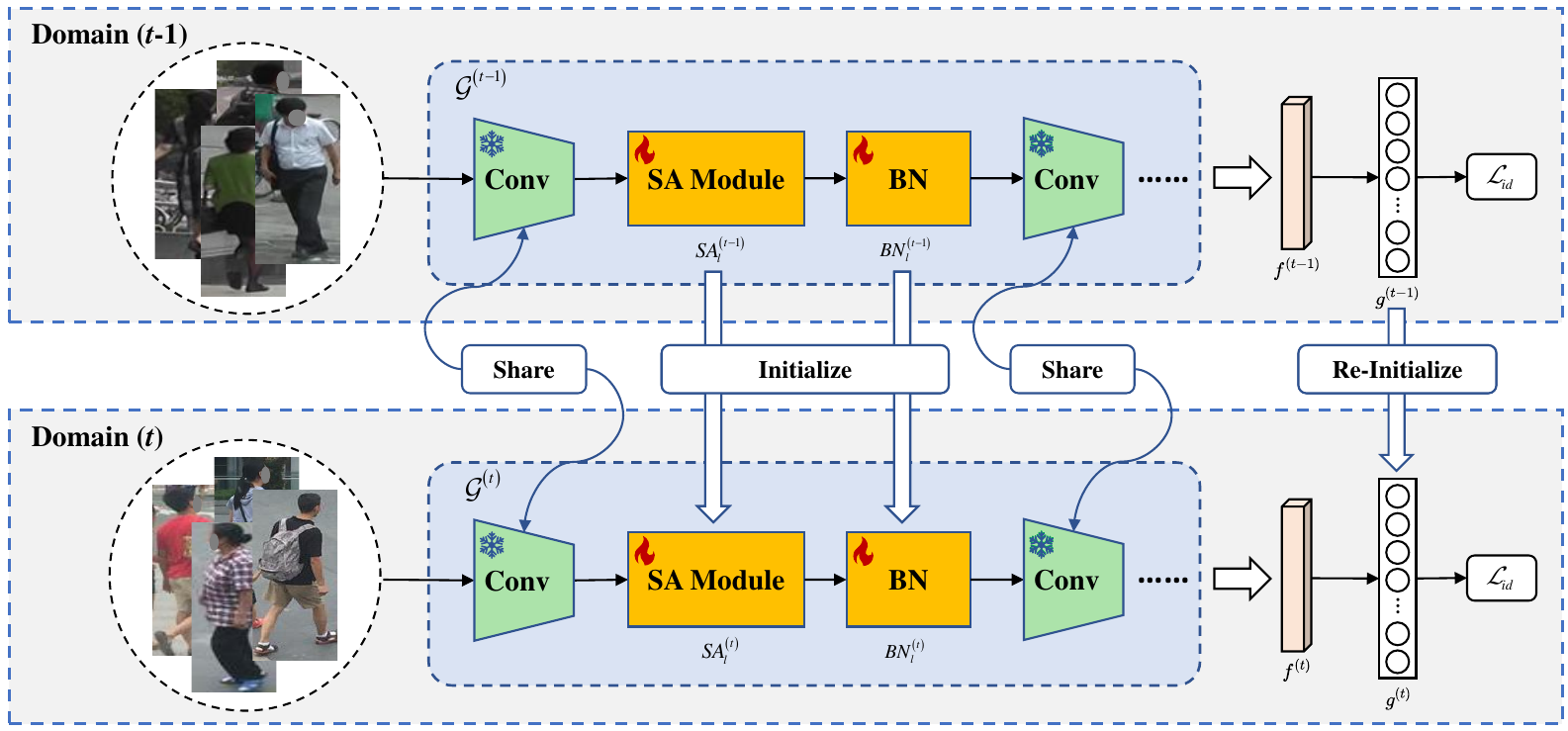}
  \caption{\textbf{The framework of DASA.} The acquired robust and general human semantics from pre-training are used as shared knowledge, which is kept in the frozen Conv layers. At each training step, we adapt the acquired knowledge from the pre-training to application domain by tuning BN layers and adopting the lightweight Semantics Adaption (SA) modules. During the lifelong evolution process, the previously learned BN and SA are used for initialization in the upcoming domain for forward knowledge transfer, while the old classifier can be discarded without increasing storage burden.}
  \label{fig:framework}
\end{figure*}

\subsection{Distribution Aligned Semantics Adaption}
With the prevalent trend of pre-training in Re-ID, we advocate taking advantage of pre-trained models for effective lifelong evolution across various domains.
In the new paradigm, we propose to utilize the acquired robust and general human semantics from pre-training as shared knowledge. At each training step, our goal is transformed to adapt the acquired semantic knowledge from the pre-training domain to the target application domain.
Two problems need to be addressed, that is, the domain distribution differences between pre-training and application, and how to effectively utilize learned human semantics. To this end, we propose to adopt efficient distribution alignment and lightweight semantics adaption, respectively.

\noindent \textbf{Distribution alignment.}
Batch Normalization (BN) as a widely-used technique in CNN can normalize the features of each domain to the same reference distribution by domain-specific normalization statistics. It has shown its advantages in regularizing models and improving their generalization ability~\cite{seo2020learning,segu2023batch}. 
Inspired by previous works, we maintain multiple sets of domain-specific BN layers to estimate the feature distribution statistics. We tune BN in each application domain so that the outputs of the updated BN layers exhibit a similar distribution to the pre-training domain, and the remaining frozen pre-trained Conv layers can receive stable input activations. In this way, the learned human semantics can be better used, which improves the discriminability of the model in each application domain.

Concretely, the data $D^{(t)}$ of each domain follows a domain-specific Gaussian distribution $N (\mu^{(t)}, {\sigma^{(t)}}^{2})$. At training time, the BN at each layer $l$ collects domain-specific batch statistics $({\mu}^{(t)}_{l}, {{\sigma}^{(t)}_{l}}^{2})$ of input feature maps, while updating the corresponding moving averages $(\overline{\mu}^{(t)}_{l}, {\overline{\sigma}^{(t)}_{l}}^{2})$ to approximate the domain distribution statistics. The calculation of BN is formulated as follows:
\begin{equation}
{\rm BN}(f^{(t)}_{l}) = {\gamma}_{l} \ \frac{f^{(t)}_{l} - {\mu}^{(t)}_{l}}{\sqrt{{{\sigma}^{(t)}_{l}}^{2} + \epsilon}} + {\beta}_{l},
\label{eq:bn}
\end{equation}
where $f^{(t)}_{l}$ is the input feature map at layer $l$, ${\gamma}_{l}$ and ${\beta}_{l}$ are learnable affine parameters for linear transformation, and $\epsilon > 0$ is a small constant to avoid the numerical problem.

After training, we can obtain the statistics and affine parameters for each domain, \textit{i.e.}, $BN^{(t)} = {\left \{ \overline{\mu}^{(t)}_{l}, {\overline{\sigma}^{(t)}_{l}}^{2}, {\gamma}_{l}, {\beta}_{l} \right \}}^{L_{BN}}_{l=1}$, where $L_{BN}$ is the number of BN layers in the CNN model.
To achieve the forward transfer of knowledge from the learned domain to the next, we use the $BN^{(t-1)}$ of the previous domain as the initial for training $BN^{(t)}$ in the new domain.

\noindent \textbf{Semantics adaption.}
Having compensated for the distribution differences between the pre-training domain and the application domain, it is time to consider how to efficiently utilize the acquired human semantic knowledge in the application domain.
The learned human semantics from pre-training are regarded as robust and general knowledge for Re-ID, so we freeze the pre-trained Conv layers to keep the shared knowledge. 
Although directly utilizing these general semantics for Re-ID has achieved decent results, it may contribute to inferior performance when the application domain shows great complexity in scenes and pedestrians (as shown in Tab.~\ref{tab:ablation}). Therefore, it is necessary to adapt the learned general semantics to specific application domains.

To this end, we introduce the lightweight Semantics Adaption (SA) module after each frozen Conv layer to aggregate and refine learned semantics effectively.
For the sake of efficiency, it could be a good choice to implement SA as one depth-wise Conv layer.
Assuming $M$ is the channel dimension of the input feature map, the depth-wise Conv layer adopts $M$ kernels and each for one channel of the input feature map, respectively.
One of the direct designs is using the 1~$\times$~1 depth-wise Conv layer. However, we find it has no effect on performance improvement and even deteriorates the discriminative ability of the model (as shown in Fig.~\ref{fig:avg_kernel_size}).
It makes sense since there is no interaction in the spatial dimension to achieve semantics adaption. Intuitively, the information interaction across spatial is vital to capture discriminative semantics, such as body figures, and improve the robustness of learned semantics in complex scenes. Therefore, we propose to apply a relatively large kernel size to aggregate and refine the general semantics.
Note that since the SA module performs convolutional operation at each channel separately, the increase in kernel size would not result in a significant increase in parameters. In our experiments, we find that the kernel size of 5~$\times$~5 can achieve a good balance between effectiveness and efficiency.

There are some differences when comparing the SA module with LoRA~\cite{hu2021lora}, which adds trainable pairs of rank decomposition matrices in parallel to existing weight matrices. Technically, LoRA updates the original weights and changes the calculation operator, adapting the model for different tasks. Differently, the SA module aims to leverage general human semantics, improving the Re-ID discriminative ability of the model in different application domains. Adapting the general human semantics to more representative identity semantics represents a process from coarse to fine, showing great ability to evolve the model for lifelong application. Incorporating reparametrization techniques like LoRA~\cite{hu2021lora} can be a prospective future work.

After training in each domain, we can obtain a set of domain-specific SA modules, \textit{i.e.}, $SA^{(t)} = {\left \{ w^{(t)}_{l} \right \}}^{L_{Conv}}_{l=1}$, where $w^{(t)}_{l}$ denotes the parameters of the SA module at layer $l$, and $L_{Conv}$ is the number of Conv layers in the CNN model. Similarly to our distribution alignment design, to achieve the forward transfer of knowledge from the learned domain to the next, we use the $SA^{(t-1)}$ as the initial to train $SA^{(t)}$ effectively.

\subsection{Overall Pipeline}
The framework is shown in Fig.~\ref{fig:framework}. At each training step, we use the basic identity classification loss $\mathcal{L}_{id}$~\cite{luo2019bag} for supervision. After training the model, we can obtain the domain-specific $BN^{(t)}$ and $SA^{(t)}$ as introduced above. Both of them are lightweight and stored for taking advantage of the learned semantics at the inference stage.
Note that the extra parameters introduced per domain are not significant compared with other competitors as shown in Fig.~\ref{fig:param_storage}. During testing, according to the camera information of the person image, corresponding $BN^{(t)}$ and $SA^{(t)}$ are used. 
With the help of $BN^{(t)}$ and $SA^{(t)}$, the model can efficiently evolve and adapt to new application domains, making it effective and practical for new deployment and wide usage in real-world scenarios.

\section{Experiments}

\subsection{Experimental Setup}
\label{subsec:exp_setup}
\noindent \textbf{Datasets.}
We evaluate our proposed method following the widely-used LReID setting~\cite{wu2021generalising,ge2022lifelong,yu2023lifelong}. Specifically, we investigate its effectiveness in two different training orders on various person Re-ID datasets.
\textbf{Order 1}: Market-1501~\cite{market1501} $\rightarrow$ DukeMTMC-reID~\cite{zheng2017unlabeled} $\rightarrow$ CUHK-SYSU~\cite{xiao2017joint} $\rightarrow$ MSMT17~\cite{wei2018person}. \textbf{Order 2}: VIPeR~\cite{gray2008viewpoint} $\rightarrow$ Market-1501 $\rightarrow$ CUHK-SYSU $\rightarrow$ MSMT17.
Note that DukeMTMC-reID is only used for academic use and fair comparison without identifying or showing pedestrian images. CUHK-SYSU is modified from the original for person search and rearranged following~\cite{wu2021generalising}. We split VIPeR into training and testing sets following \cite{yu2023lifelong}. For other datasets, we follow their original training and evaluation protocols.
To investigate its generalization capability, we also evaluate it on the unseen datasets, including CUHK01~\cite{li2013human}, CUHK02~\cite{li2013locally}, GRID~\cite{loy2010time}, SenseReID~\cite{zhao2017spindle}, PRID~\cite{hirzer2011person}.

We also evaluate the influence of using different datasets for pre-training, including ImageNet~\cite{deng2009imagenet} for image classification, LUPerson~\cite{fu2021unsupervised} and LUPerson-NL~\cite{fu2022large} for person Re-ID. 
LUPerson is a large-scale unlabeled dataset of 4M images of over 200K identities. LUPerson-NL is derived from LUPerson by applying an online multi-object tracking system on the raw videos of LUPerson. LUPerson-NL consists of 10M images with about 430K identities collected from 21K scenes. 
Unless otherwise specified, we use the LUPerson-NL pre-trained weights for our method, which can provide robust and general human semantics as shared knowledge for lifelong evolution.

\noindent \textbf{Evaluation metrics.}
Mean Average Precision (mAP) and Rank-1 accuracy (R-1) are adopted for performance evaluation on seen domains after each training step. We also report average accuracies by averaging mAP and R-1 on all seen domains (denoted as $\overline{s}_{\rm mAP}$ and $\overline{s}_{\rm R-1}$), as well as performance on unseen domains after the last training step.

\subsection{Implementation Details}
We adopt ResNet-50~\cite{resnet} as the backbone model following previous works~\cite{pu2021lifelong,wu2021generalising,ge2022lifelong,yu2023lifelong}, and the input images are resized to $256 \times 128$ with random horizontal flipping, padding, random cropping, and random erasing~\cite{zhong2020random} for data augmentation. The batch size is set to 128, with 2 samples per pedestrian. 
Adam optimizer~\cite{kingma2014adam} with weight decay of $5 \times 10^{-4}$ is adopted, with the warmup strategy that linearly increases the learning rate from $3.5 \times 10^{-6}$ to $3.5 \times 10^{-4}$ in the first 10 epochs. The learning rate is decreased by a factor of 10 at the 30th epoch for the first dataset and at the 10th epoch for other datasets. Each dataset is trained for 80 epochs.
For the SA module, the kernel size is set to 5 $\times$ 5 with stride 1, and the bias term is removed for the sake of the number of parameters and its limited effect.

\begin{table*}[t]
  \centering
  \caption{\textbf{Comparison with the state-of-the-art methods in the LReID setting of Order 1.}
  ``\textit{w/} Ex.'' denotes rehearsal-based methods using exemplars. The results are reported after the last training step. The best results are shown in bold. }
  \label{tab:compare_SOTA}
  \resizebox{1\linewidth}{!}{
    \begin{tabular}{lcccccccccccc}
    \toprule
    \multirow{2}[0]{*}{Methods} & \multirow{2}[0]{*}{Reference} & \multirow{2}[0]{*}{\textit{w/} Ex.}
    & \multicolumn{2}{c}{Market-1501}  & \multicolumn{2}{c}{DukeMTMC} & \multicolumn{2}{c}{CUHK-SYSU} & \multicolumn{2}{c}{MSMT17} & \multicolumn{2}{c}{Average}  \\
    \cmidrule(r){4-5} \cmidrule(r){6-7} \cmidrule(r){8-9} \cmidrule(r){10-11} \cmidrule(r){12-13}
    & & & mAP & R-1 & mAP & R-1 & mAP & R-1 & mAP & R-1 & $\overline{s}_{\rm mAP}$ & $\overline{s}_{\rm R-1}$ \\
    \midrule

    {LwF~\cite{li2017learning}} & TPAMI17 & & 72.4 & 88.3 & 59.3 & 74.6 & 86.1 & 87.2 & 36.1 & 63.0 & 63.5 & 78.3 \\
    {SPD~\cite{tung2019similarity}} & ICCV19 & & 71.4 & 87.8 & 58.5 & 75.1 & 86.5 & 88.0 & 37.7 & 64.3 & 63.4 & 78.9 \\
    {CRL~\cite{zhao2021continual}} & WACV21 & & 60.5 & 83.7 & 51.9 & 71.8 & 83.7 & 86.4 & 41.2 & 67.1 & 59.3 & 77.3 \\
    {BiC~\cite{wu2019large}} & CVPR19 & $\checkmark$ & 53.4 & 75.8 & 37.7 & 55.4 & 81.4 & 84.2 & 13.1 & 33.0 & 46.4 & 62.1 \\
    {WA~\cite{zhao2020maintaining}} & CVPR20 & $\checkmark$ & 52.1 & 73.2 & 36.0 & 54.6 & 81.1 & 83.7 & 17.2 & 39.6 & 46.6 & 62.8 \\
    \midrule

    {AKA~\cite{pu2021lifelong}} & CVPR21 & & 59.7 & 80.1 & 32.7 & 48.3 & 82.0 & 84.4 & 17.1 & 34.9 & 47.9 & 61.9 \\
    {GwFReID~\cite{wu2021generalising}} & AAAI21 & $\checkmark$ & 60.9 & 81.6 & 46.7 & 66.5 & 81.4 & 83.9 & 25.9 & 52.4 & 53.7 & 71.1 \\
    {KRC~\cite{yu2023lifelong}} & AAAI23 & $\checkmark$ & 60.3 & 82.3 & 58.7 & 72.7 & 88.9 & 90.5 & 43.3 & 67.7 & 62.8 & 78.3 \\
    {PTKP~\cite{ge2022lifelong}} & AAAI22 & $\checkmark$ & 75.8 & 89.7 & 62.0 & 76.7 & 85.0 & 86.3 & 34.5 & 60.9 & 64.3 & 78.4 \\
    \midrule

    {DASA} & {Ours} &  & \textbf{86.2} & \textbf{94.6} & \textbf{77.3} & \textbf{87.6} & \textbf{93.8} & \textbf{94.7} & \textbf{49.3} & \textbf{73.6} & \textbf{76.7} & \textbf{87.6} \\
    \bottomrule
    \end{tabular}}
\end{table*}

\begin{figure*}[t]
  \centering
  \includegraphics[width=\linewidth]{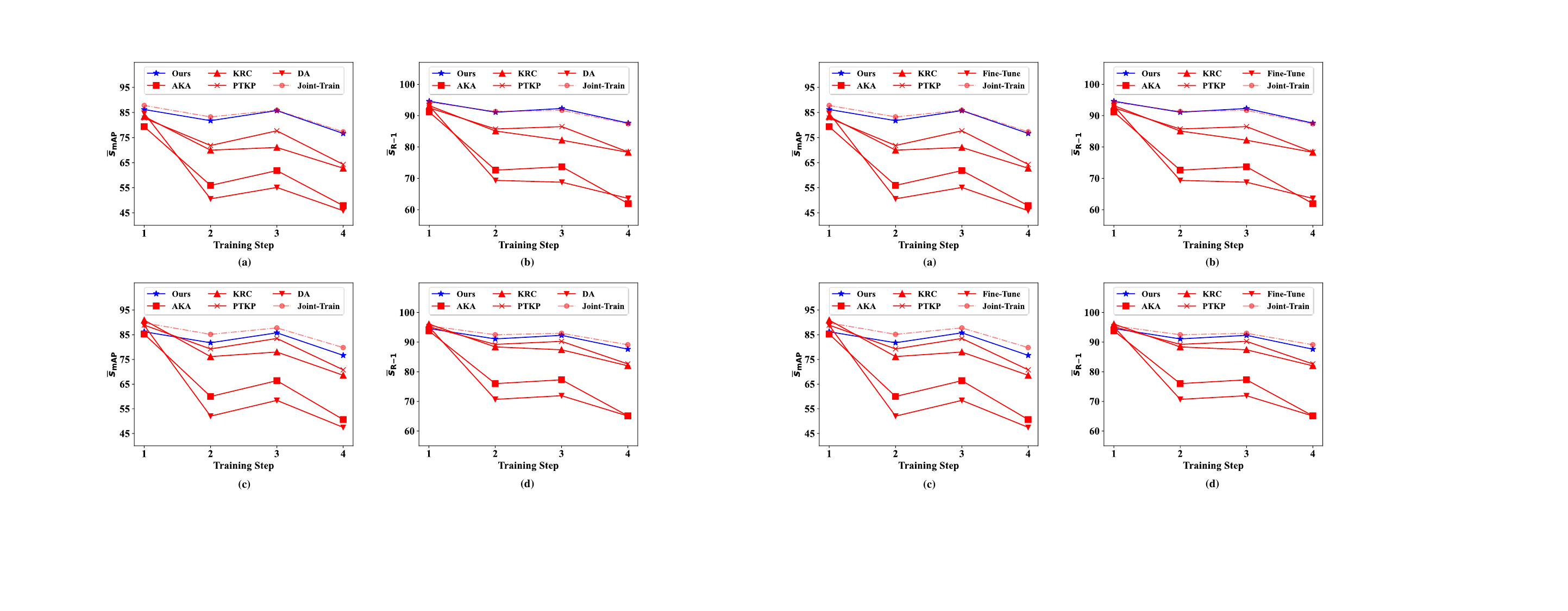}
  \caption{\textbf{Comparison of average accuracies at different training steps of Order~1 w.r.t. (a) $\overline{s}_{\rm mAP}$ and (b) $\overline{s}_{\rm R-1}$.}}
  \label{fig:avg_seen_mAP_rank1}
\end{figure*}

\subsection{Comparison with State-of-the-Arts}
\noindent \textbf{Results on seen domains.}
We compare our proposed method with advanced lifelong learning methods for classification and recognition~\cite{li2017learning,tung2019similarity,zhao2021continual,wu2019large,zhao2020maintaining,rebuffi2017icarl,hou2019learning}, and those specially designed for LReID~\cite{pu2021lifelong,wu2021generalising,yu2023lifelong,ge2022lifelong}. The results of training Orders 1 and 2 are shown in Tabs.~\ref{tab:compare_SOTA} and~\ref{tab:compare_SOTA2}, respectively. Comparison methods for LReID are reproduced in the same experimental environment according to their released code and training settings.
Fine-Tune denotes fine-tuning on different datasets sequentially. Joint-Train denotes combining all datasets for joint training.
As shown in Tab.~\ref{tab:compare_SOTA}, due to the discrepancies across domains, although rehearsal-based methods show great performance in the classification task~\cite{wu2019large,zhao2020maintaining}, it contributes to inferior performance gain when being applied to LReID. 
For the methods designed for LReID, AKA~\cite{pu2021lifelong} struggles to maintain a balance between forward knowledge learning and retention. Other advanced methods~\cite{ge2022lifelong,yu2023lifelong,wu2021generalising} draw support from exemplars following the knowledge distillation paradigm. 
Without relying on impractical exemplars, DASA still outperforms them significantly. 

The previously acquired knowledge will not be forgotten in our LReID paradigm. However, previous methods forget more and more with the increasing of subsequent domains. Previous studies~\cite{pu2021lifelong,ge2022lifelong} have discussed the forgetting phenomenon of the model on seen datasets. Since our LReID paradigm avoids forgetting, we compare the average accuracies at different training steps, which more comprehensively evaluates the capabilities in knowledge preservation and knowledge updating, both of which are essential for LReID. As shown in Fig.~\ref{fig:avg_seen_mAP_rank1}, DASA achieves better average accuracies during the lifelong evolution process on the four datasets. The results show its effectiveness in adapting to the application domains and acquiring knowledge.

\begin{table*}[t]
  \centering
  \caption{\textbf{Comparison with the state-of-the-art methods in the LReID setting of Order 2.} ``\textit{w/} Ex.'' denotes rehearsal-based methods using exemplars. The results are reported after the last training step.
  The best results are shown in bold. 
  }
  \label{tab:compare_SOTA2}
  \resizebox{1\linewidth}{!}{
    \begin{tabular}{lcccccccccccc}
    \toprule
    \multirow{2}[0]{*}{Methods} & \multirow{2}[0]{*}{Reference} & \multirow{2}[0]{*}{\textit{w/} Ex.}
    & \multicolumn{2}{c}{VIPeR}  & \multicolumn{2}{c}{Market-1501} & \multicolumn{2}{c}{CUHK-SYSU} & \multicolumn{2}{c}{MSMT17} & \multicolumn{2}{c}{Average}  \\
    \cmidrule(r){4-5} \cmidrule(r){6-7} \cmidrule(r){8-9} \cmidrule(r){10-11} \cmidrule(r){12-13}
    & & & mAP & R-1 & mAP & R-1 & mAP & R-1 & mAP & R-1 & $\overline{s}_{\rm mAP}$ & $\overline{s}_{\rm R-1}$ \\
    \midrule
    {LwF~\cite{li2017learning}} & TPAMI17 & & 54.6 & 45.3 & 53.9 & 74.6 & 81.1 & 83.7 & 14.6 & 32.3 & 51.1 & 59.0 \\
    {iCarL~\cite{rebuffi2017icarl}} & CVPR17 & $\checkmark$ & 67.0 & 56.6 & 56.9 & 78.7 & 80.0 & 82.9 & 10.2 & 24.1 & 53.6 & 60.6 \\
    {UCIR~\cite{hou2019learning}} & CVPR19 & $\checkmark$ & 66.8 & 56.3 & 45.5 & 65.9 & 68.8 & 70.5 & 12.8 & 29.3 & 48.5 & 55.5 \\
    {BiC~\cite{wu2019large}} & CVPR19 & $\checkmark$ & 61.2 & 50.4 & 47.4 & 68.9 & 71.6 & 22.4 & 72.5 & 42.1 & 50.7 & 58.5 \\
    {WA~\cite{zhao2020maintaining}} & CVPR20 & $\checkmark$ & 58.0 & 48.1 & 50.9 & 70.3 & 70.4 & 71.9 & 18.6 & 38.8 & 49.5 & 57.2 \\
    \midrule
    {AKA~\cite{pu2021lifelong}} & CVPR21 & & 61.7 & 50.6 & 28.3 & 50.7 & 76.9 & 79.6 & 13.4 & 28.0 & 45.1 & 52.2 \\
    {PTKP~\cite{ge2022lifelong}} & AAAI22 & $\checkmark$ & 59.5 & 45.9 & 49.3 & 69.9 & 71.4 & 73.4 & 18.0 & 37.7 & 49.6 & 56.7 \\
    {KRC~\cite{yu2023lifelong}} & AAAI23 & $\checkmark$ & 76.7 & 66.5 & 65.7 & 84.7 & 88.6 & 90.3 & 42.2 & 66.2 & 68.3 & 76.9 \\
    \midrule
    
    {DASA} & Ours &  & \textbf{82.2} & \textbf{73.7} & \textbf{84.9} & \textbf{93.9} & \textbf{92.3} & \textbf{93.3} & \textbf{45.2} & \textbf{70.7} & \textbf{76.2} & \textbf{82.9} \\
    \bottomrule
    \end{tabular}}
\end{table*}

As shown in Tab.~\ref{tab:compare_SOTA2}, our proposed DASA also achieves state-of-the-art results when training in Order 2. On the small-scale VIPeR~\cite{gray2008viewpoint} dataset with only 632 person images for training, DASA still shows great superiority.
By comparing the results of Orders 1 and 2, we have the following observations. 
\textbf{(1)} The training order has a great impact on the state-of-the-art competitors. For example, PTKP~\cite{ge2022lifelong} achieves slightly better results than KRC~\cite{yu2023lifelong} in Order 1, but much worse results in Order 2. 
Another example is that on Market-1501, PTKP shows much better results in Order 1 compared with Order 2 (surpassing about 20\% on mAP and R-1). 
\textbf{(2)} KRC performs better than PTKP on the small-scale VIPeR. In contrast, affected by the exemplars from VIPeR, PTKP achieves bad incremental accuracies on the following datasets. These results show the great sensitivity of advanced LReID methods to the chosen training datasets. 
\textbf{(3)} The catastrophic forgetting problem is not well addressed by other advanced methods. For instance, KRC performs worse on Market-1501 in Order 1 since more new domains participate in the lifelong learning process than those in Order 2. In contrast, DASA is robust to different training orders and datasets.

\begin{figure}[t]
\begin{minipage}[t]{0.5\linewidth}
  \centering
  \vspace{0pt}
  \includegraphics[width=0.8\linewidth]{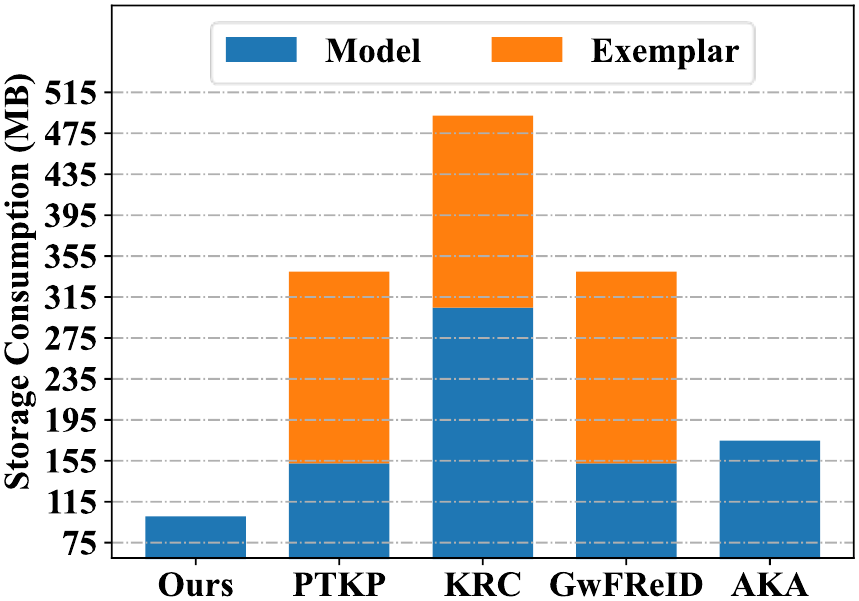}
  \caption{\textbf{Comparison of storage consumption for LReID methods.} Results are calculated after the last training step.
  }
  \label{fig:param_storage}
\end{minipage} \hfill
\begin{minipage}[t]{0.49\linewidth}
    \centering
    \scriptsize
    \vspace{0pt}
    \captionof{table}{\textbf{Storage consumption of different components.} Results are calculated after the last training step.}
    \label{tab:detailed_param_storage}
    \setlength{\tabcolsep}{3.5mm}{
        \begin{tabular}{lr}
        \toprule
        \multicolumn{1}{c}{\textbf{Component}} & \textbf{Storage (MB)} \\
        \midrule
        Backbone & 89.684 \\
        \midrule
        Classifier & 62.703 \\
        Exemplars & 187.500 \\
        \midrule
        $BN^{(t)}$ & 0.210 \\
        $SA^{(t)}$ & 2.533 \\
        \bottomrule
        \end{tabular}}
\end{minipage}
\end{figure}

\noindent \textbf{Comparison of storage consumption.}
Storage consumption is also crucial in lifelong scenarios, otherwise, we can save all previous models to prevent catastrophic forgetting. We compare the storage consumption for different LReID methods after the last training step (Order 1) in Fig.~\ref{fig:param_storage} and Tab.~\ref{tab:detailed_param_storage}.
All competitors~\cite{ge2022lifelong,yu2023lifelong,wu2021generalising} have to store the classifier for knowledge distillation to relieve forgetting. As introduced in Sec.~\ref{subsec:knowledge_distillation}, as the process of lifelong learning, the dimension of the identity category increases dramatically. The classifier, implemented as one fully connected layer, would consume more and more, whose size is almost seven-tenths of the size of the model after the last training step as shown in Tab.~\ref{tab:detailed_param_storage}. What's worse, an old model having the same large classifier is used for knowledge distillation, further exacerbating storage consumption.

In addition, they all draw support from exemplars. They keep 250 identities with 2 person images per identity at each training step. However, on the one hand, due to privacy issues, saving exemplars is unrealistic. On the other hand, due to the open categories of pedestrians, saving a large number of identity images can consume storage a lot. The storage consumption of exemplars even surpasses that of both the backbone model and the classifier.
Note that in GwFReID~\cite{wu2021generalising}, all identities are selected into the exemplar pool, but when comparing it in Fig.~\ref{fig:param_storage}, we still assume it saves 250 identities per step, otherwise its storage cost would be enormous (about 1578MB for exemplars).

However, our proposed method is free of exemplars. At each step, only a more lightweight classifier is needed for identity classification during training. Once the training is finished, the classifier can also be discarded freely and would not affect the lifelong evolution of the model.
As shown in Tab.~\ref{tab:detailed_param_storage}, the introduced storage consumption at each training step is only equal to 3\% of the backbone model (\textit{v.s.}, 70\% on average for advanced competitors). All the results indicate the significant advantages of our proposed DASA in terms of efficiency and effectiveness in LReID.

\noindent \textbf{Results on unseen domains.}
Our proposed DASA shows great anti-forgetting capability and outstanding performance in application domains. We also investigate its generalization capability on unseen domains. We follow the competitors to use the model at the last training step which accumulates rich knowledge from all seen domains. 
As shown in Tabs.~\ref{tab:unseen_order1} and \ref{tab:unseen_order2}, the generalization ability of DASA outperforms that of all other advanced methods as well as Fine-Tune and Joint-Train on all datasets in both training orders. The results indicate that robust and general human semantic knowledge is useful for generalization and DASA shows great effectiveness in generalizing acquired knowledge.

\begin{table}[t]
\begin{minipage}[t]{0.49\linewidth}
  \centering
  \scriptsize
  \vspace{0pt}
  \caption{\textbf{Comparison with state-of-the-art methods on unseen domains after training in Order 1.} The results (Rank-1) are given by the trained model after the last training step.}
  \label{tab:unseen_order1}
  \setlength{\tabcolsep}{0.49mm}{
    \begin{tabular}{lccc}
    \toprule
    \textbf{Methods} & \textbf{CUHK01} & \textbf{GRID} & \textbf{SenseReID} \\
    \midrule
    Fine-Tune & 55.7 & 7.4 & 28.7 \\
    Joint-Train & 66.0 & 19.4 & 34.9 \\
    \midrule
    LwF~\cite{li2017learning} & 73.0 & 20.1 & 39.1 \\
    SPD~\cite{tung2019similarity} & 73.9 & 20.8 & 42.1 \\
    CRL~\cite{zhao2021continual} & 71.4 & 18.7 & 40.1 \\
    PTKP~\cite{ge2022lifelong} & 69.4 & 25.3 & 46.1 \\
    KRC~\cite{yu2023lifelong} & 78.0 & 25.5 & 44.0 \\
    \midrule
    DASA & \textbf{83.5} & \textbf{36.0} & \textbf{50.0} \\
    \bottomrule
    \end{tabular}}
\end{minipage} \hfill
\begin{minipage}[t]{0.49\linewidth}
  \centering
  \scriptsize
  \vspace{0pt}
  \caption{\textbf{Comparison with state-of-the-art methods on unseen domains after training in Order 2.} The results (mAP) are given by the trained model after the last training step.}
  \label{tab:unseen_order2}
  \vspace{-0.08in}
  \setlength{\tabcolsep}{0.8mm}{
    \begin{tabular}{lcccc}
    \toprule
    \textbf{Methods} & \textbf{CUHK01} & \textbf{CUHK02} & \textbf{PRID} \\
    \midrule
    Fine-Tune & 48.7 & 40.5 & 11.4 \\
    Joint-Train & 65.2 & 56.8 & 14.8 \\
    \midrule
    LwF~\cite{li2017learning} & 60.1 & 52.0 & 27.0 \\
    iCarL~\cite{rebuffi2017icarl} & 60.4 & 55.1 & 34.9 \\
    BiC~\cite{wu2019large} & 50.4 & 44.6 & 24.1 \\
    WA~\cite{zhao2020maintaining} & 50.2 & 48.5 & 21.9 \\
    PTKP~\cite{ge2022lifelong} & 47.7 & 47.6 & 24.2 \\
    KRC~\cite{yu2023lifelong} & 76.6 & 66.9 & 49.1 \\
    \midrule
    DASA & \textbf{78.6} & \textbf{70.2} & \textbf{64.3} \\
    \bottomrule
    \end{tabular}}
\end{minipage}
\end{table}

\subsection{Ablation Studies}
\begin{figure*}[t]
\centering
  \includegraphics[width=\linewidth]{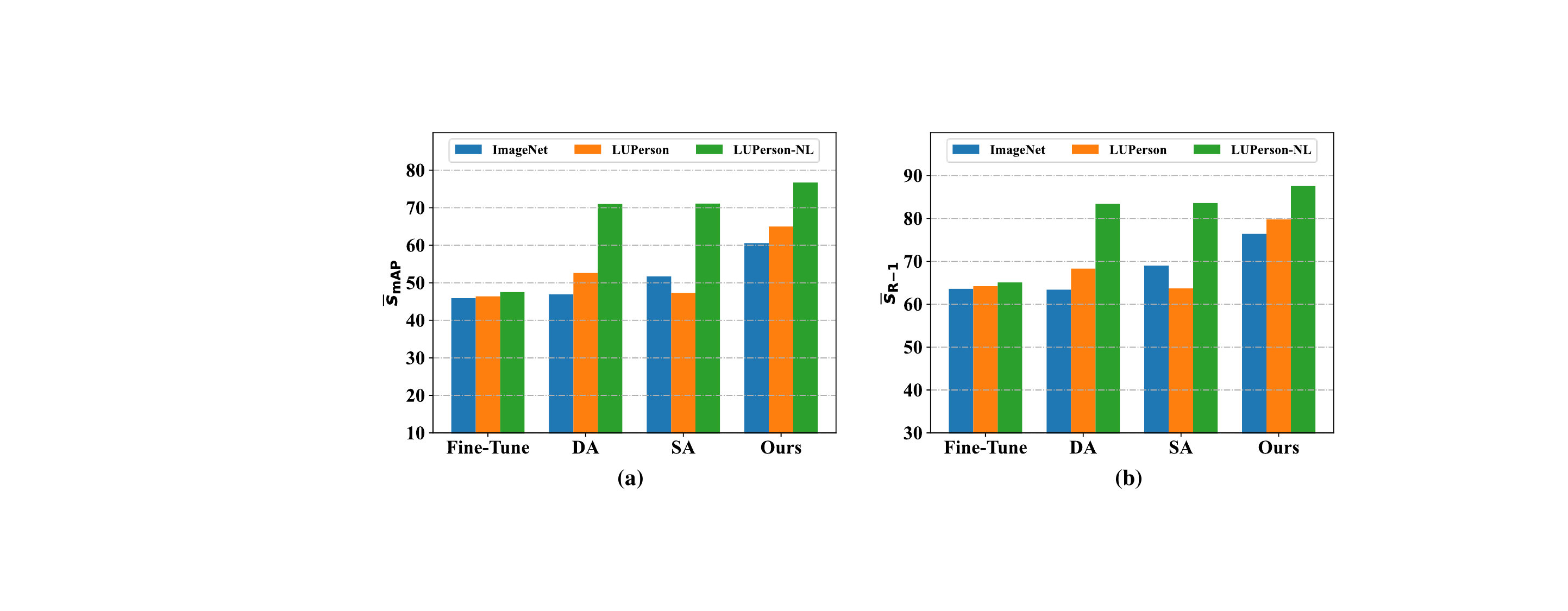}
  \caption{\textbf{Influence of different pre-training data for LReID.} (a) $\overline{s}_{\rm mAP}$ and (b) $\overline{s}_{\rm R-1}$ at the last training step are reported using different methods.
  }
  \label{fig:pretrain_mAP_rank1}
\end{figure*}

For a fairer comparison and to validate our designs, we conduct all ablation experiments using the LUPerson-NL~\cite{fu2022large} pre-trained weights in Tab.~\ref{tab:ablation}. Without loss of generality, ablation experiments are conducted in Order 1 by default.
Besides state-of-the-art methods, we also compare (1) Fine-Tune: fine-tuning the pre-trained model on different datasets sequentially; 
(2) DA: tuning BN layers to achieve distribution alignment at each training step; 
(3) SA: freezing the pre-trained model and only adding and tuning our proposed SA modules; 
(4) our proposed DASA; 
(5) Joint-Train: combining all datasets for joint training.
Note that for SA, since BNNeck~\cite{luo2019bag} is not trained during pre-training, it is also trained at each training step.
In Fig.~\ref{fig:pretrain_mAP_rank1}, we further investigate the influence of various pre-training choices on different methods. 
We also explore the choices of kernel sizes for the SA module, the positions to use SA, and the effectiveness of SA in adapting semantics via attention map visualization.

\begin{table*}[t]
  \centering
  \caption{\textbf{Ablation Studies of our method.} \textit{All methods use LUPerson-NL pre-trained weights.} The results are reported after the last training step in Order 1. ``$*$" denotes adding the SA module to the original ResNet-50.}
  \label{tab:ablation}
  \resizebox{0.96\linewidth}{!}{
    \begin{tabular}{lcccccccccc}
    \toprule
    \multirow{2}[0]{*}{Methods} & \multicolumn{2}{c}{Market-1501} & \multicolumn{2}{c}{DukeMTMC} & \multicolumn{2}{c}{CUHK-SYSU} & \multicolumn{2}{c}{MSMT17} & \multicolumn{2}{c}{Average}  \\
    \cmidrule(r){2-3} \cmidrule(r){4-5} \cmidrule(r){6-7} \cmidrule(r){8-9} \cmidrule(r){10-11}
    & mAP & R-1 & mAP & R-1 & mAP & R-1 & mAP & R-1 & $\overline{s}_{\rm mAP}$ & $\overline{s}_{\rm R-1}$ \\
    \midrule

    {AKA~\cite{pu2021lifelong}} & 66.7 & 83.7 & 38.3 & 53.3 & 84.5 & 86.1 & 19.1 & 37.4 & 52.2 & 65.1 \\
    {KRC~\cite{yu2023lifelong}} & 68.4 & 85.8 & 64.8 & 76.6 & 91.9 & 93.5 & 49.2 & 72.2 & 68.6 & 82.0 \\
    {PTKP~\cite{ge2022lifelong}} & 83.6 & 93.3 & 69.1 & 81.3 & 89.4 & 90.4 & 41.1 & 65.6 & 70.8 & 82.7 \\
    \midrule

    Fine-Tune$^{*}$ & 25.9 & 50.6 & 31.1 & 49.4 & 65.9 & 68.9 & 36.9 & 62.0 & 40.0 & 57.7 \\
    {Fine-Tune} & 30.0 & 56.3 & 38.7 & 55.6 & 76.4 & 79.2 & 44.9 & 69.2 & 47.5 & 65.1 \\
    {DA} & 80.4 & 91.6 & 70.9 & 83.5 & 93.0 & 94.1 & 39.8 & 64.5 & 71.0 & 83.4 \\
    {SA} & 80.1 & 92.1 & 72.4 & 84.2 & 91.8 & 93.0 & 40.1 & 65.2 & 71.1 & 83.6 \\
    \midrule
    {DASA (Ours)} & \textbf{86.2} & \textbf{94.6} & \textbf{77.3} & \textbf{87.6} & \textbf{93.8} & \textbf{94.7} & \textbf{49.3} & \textbf{73.6} & \textbf{76.7} & \textbf{87.6} \\
    \midrule
    \rowcolor{gray!20} {Joint-Train} & 89.8 & 95.5 & 80.5 & 89.5 & 93.0 & 94.0 & 55.9 & 77.5 & 79.8 & 89.1 \\
    \rowcolor{gray!20} Joint-Train$^{*}$ & 90.2 & 95.7 & 81.6 & 90.3 & 93.1 & 94.2 & 57.2 & 78.0 & 80.5 & 89.6 \\
    \bottomrule
    \end{tabular}}
\end{table*}

\noindent \textbf{Effectiveness of distribution alignment.}
As shown in Tab.~\ref{tab:ablation}, DA can achieve better results than advanced competitors~\cite{pu2021lifelong,yu2023lifelong,ge2022lifelong}, confirming the great impact of data distribution on LReID. Tuning BN layers eliminates the distribution differences between pre-training and application, which contributes to taking advantage of learned human semantics effectively. Unfortunately, conventional architecture-based lifelong learning methods designed for classification or recognition tasks~\cite{yoon2018lifelong,mallya2018packnet} do not concern the influence of BN layers. They try to keep and use acquired knowledge but do not mitigate the domain distribution gap, making them unsuitable for LReID.

\noindent \textbf{Effectiveness of the SA module.}
The results in Tab.~\ref{tab:ablation} show that SA also significantly beats all advanced competitors~\cite{pu2021lifelong,yu2023lifelong,ge2022lifelong} using the same pre-trained weights. It demonstrates that just using the pre-trained weights from large-scale person datasets for model initialization cannot effectively promote LReID. The knowledge distillation paradigm limits their performance upper bound. However, our proposed SA is effective in leveraging human semantics learned from pre-training to promote LReID.
We also try to add the SA module to the ResNet-50 backbone. For Fine-Tune, the previously learned semantics are destroyed and tuned to fit the current domain, so incorporating the SA module may lead to inflexible knowledge updating and severe knowledge forgetting, resulting in performance degradation. 
In contrast, SA shows its great effectiveness in our proposed LReID paradigm. Additionally, since Joint-Train can acquire all data from all application domains, the SA module can also help improve semantics with few parameters. With proper design and usage, SA can bring great benefits.

\noindent \textbf{Necessity of adopting both distribution alignment and semantics adaption.}
As shown in Tab.~\ref{tab:ablation}, combining both DA and SA achieves the best results. 
On the one hand, although the pre-trained model can acquire human semantic knowledge to represent pedestrians, the data distribution discrepancy between pre-training and application would result in ineffective use of the learned semantics for specific applications.
On the other hand, with the help of the SA module for aggregating semantic representations after distribution alignment, on the challenging DukeMTMC-reID and MSMT17 datasets, DASA brings 6.4\% and 9.5\% mAP improvement over DA, respectively.
Note that DASA achieves competitive results with Joint-Train, and even better mAP and R-1 on the CUHK-SYSU dataset. We guess it is because the image styles and data distributions of CUHK-SYSU differ greatly from others. When trained jointly, the model has to make a compromise between datasets. On the contrary, without being influenced by distribution discrepancy, DASA leverages acquired knowledge of human semantics with the SA module to represent pedestrians effectively.

\begin{figure}[t]
\begin{minipage}[t]{0.43\linewidth}
  \centering
  \vspace{0pt}
  \includegraphics[width=\linewidth]{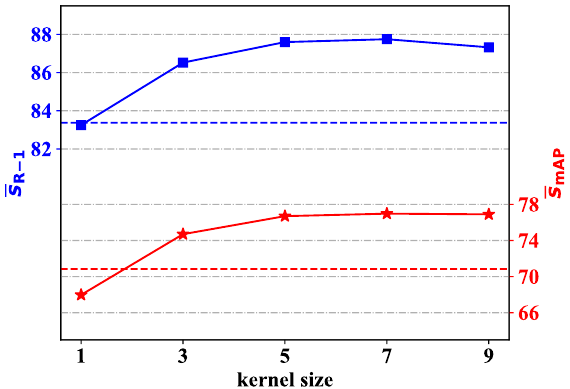}
  \caption{\textbf{Ablation Study of the kernel size for the SA module.} The average accuracies of all datasets at the last training step are reported.
  }
  \label{fig:avg_kernel_size}
\end{minipage} \hfill
\begin{minipage}[t]{0.53\linewidth}
  \centering
  \vspace{0pt}
  \captionof{table}{\textbf{Influence of applying the SA module in different layers of ResNet-50.} The average incremental accuracies of four datasets at the last training step are reported.
  }
  \label{tab:position_SA}
  \resizebox{1\linewidth}{!}{
    \scriptsize
    \begin{tabular}{ccccccc}
    \toprule
    {\multirow{2}{*}{\textbf{Methods}}} & \multicolumn{4}{c}{\textbf{Layers}} & \multirow{2}{*}{$\overline{s}_{\rm \textbf{mAP}}$} & \multirow{2}{*}{$\overline{s}_{\rm \textbf{R-1}}$} \\ 
    \cmidrule{2-5}
    & \textbf{1} & \textbf{2} & \textbf{3} & \textbf{4} & &  \\
    \midrule
    1 & & & & & 71.0 & 83.4 \\
    2 & $\checkmark$ & $\checkmark$ & & & 72.2 & 84.6 \\
    3 &  & $\checkmark$ & $\checkmark$ & & 75.3 & 86.8 \\
    4 &  &  & $\checkmark$  & $\checkmark$ & 76.0 & 87.0 \\
    5 &  & $\checkmark$ & $\checkmark$ & $\checkmark$ & 76.4 & 87.5 \\
    \midrule
    Ours & $\checkmark$ & $\checkmark$ & $\checkmark$ & $\checkmark$ & \textbf{76.7} & \textbf{87.6} \\
    \bottomrule
    \end{tabular}}
\end{minipage}
\end{figure}

\noindent \textbf{Influence of different kernel sizes of the SA module.}
In Fig.~\ref{fig:avg_kernel_size}, when setting the kernel size of the SA module to 1, we observe the performance decrease. The results support our motivation that since the 1$\times$1 depth-wise convolutional layer does not involve the spatial interaction of input feature maps technically, it cannot aggregate and adapt acquired semantic knowledge. When increasing the kernel size to 3 or 5, we can observe a significant performance improvement. However, continuing to increase the kernel size results in less performance gain. Note that setting the kernel size to 7 requires almost twice as many parameters as 5. 
Considering the performance and the number of parameters, we set the kernel size of the SA module to 5.

\noindent \textbf{Where to use the SA module.}
Considering grid search on each layer takes a great effort, we explore adopting SA in different layers of ResNet-50~\cite{resnet} in Tab.~\ref{tab:position_SA}.
\textbf{(1)} By comparing Methods 2$\sim$4, we find that the SA module brings a better effect in deep layers. It is consistent with our motivation since the deep layers are more specialized in semantics while the shallow layers are better at capturing curves and image patterns.
\textbf{(2)} Methods 2$\sim$4 all outperform Method 1, showing the effectiveness of SA in adapting acquired knowledge for better pedestrian representations.
\textbf{(3)} Method 5 shows competitive performance with Ours with even less resource cost. Considering the better performance, we apply the lightweight SA module after each frozen Conv layer. It would be also effective to adopt it only in the deep layers for more efficient deployment and application.

\noindent \textbf{Influence of different pre-training data.}
According to the results of Fig.~\ref{fig:pretrain_mAP_rank1}, we observe that \textbf{(1)} Methods based on ImageNet pre-training generally achieve inferior results. Since ImageNet includes all kinds of objects, pre-training on it brings less useful human semantics for LReID. 
\textbf{(2)} DA using LUPerson pre-trained weights outperforms that using ImageNet pre-trained weights, showing that pre-training on the person dataset provides useful knowledge for LReID. However, it still performs worse than that using LUPerson-NL pre-trained weights. The results demonstrate that when the person dataset used for pre-training is not large-scale, the model cannot acquire robust and general human semantics effectively.
\textbf{(3)} SA outperforms DA when using ImageNet pre-trained weights. On the one hand, the model does not acquire useful knowledge for LReID, since it is still suboptimal despite aligning the data distribution via DA. On the other hand, the SA module can aggregate acquired knowledge for object classification to promote LReID to a certain extent, demonstrating its effectiveness.
\textbf{(4)} DA achieves competitive or even better results with SA when using LUPerson or LUPerson-NL pre-trained weights, showing the nonnegligible impact of the data distribution gap.
\textbf{(5)} Aligning distribution or adopting SA can facilitate LReID in different pre-training choices. Combining them shows great superiority over Fine-Tune, and good pre-trained weights can expand the upper bound of DASA. It is appealing to promote LReID with our paradigm following the trend of large-scale pre-training.

\begin{figure*}[t]
\centering
  \includegraphics[width=\linewidth]{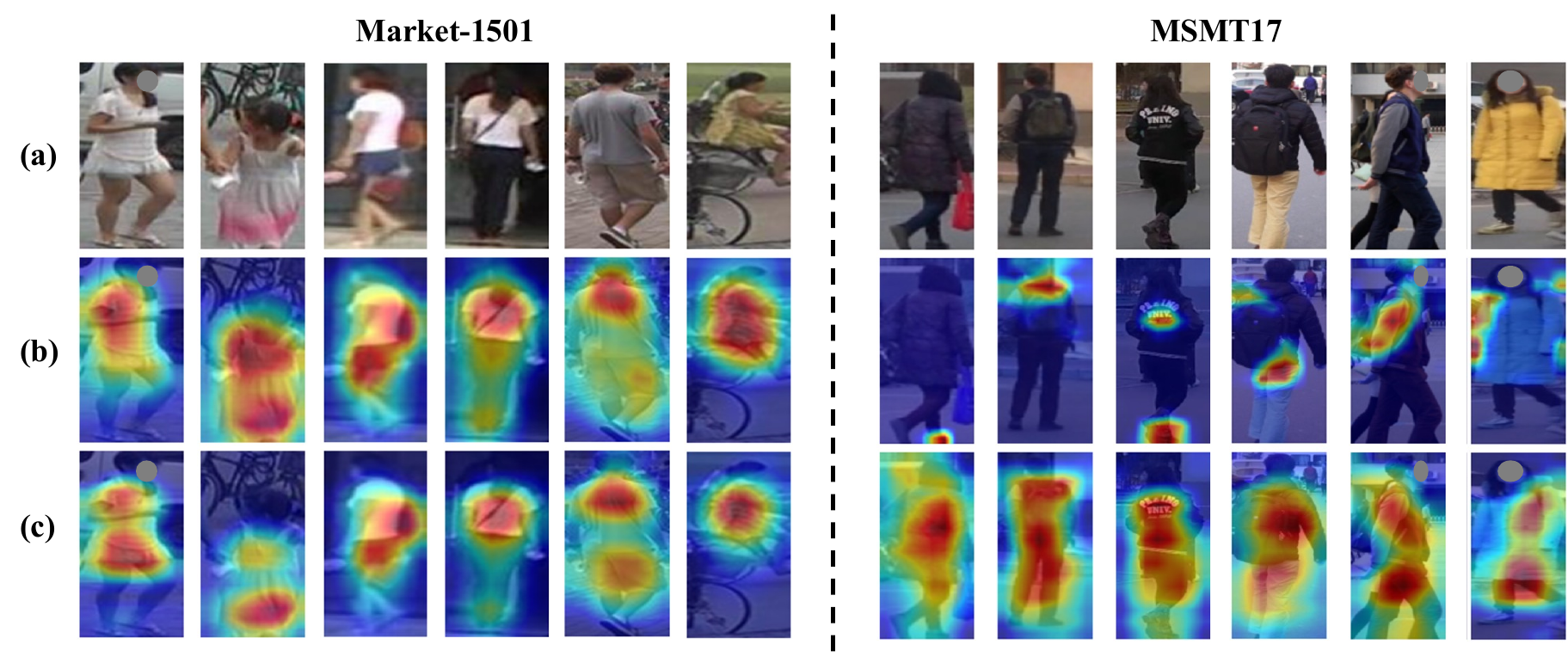}
  \caption{\textbf{Visualization of attention maps on the Market-1501 and MSMT17 datasets.} (a) Original images, (b) \textit{w/o} SA, and (c) \textit{w/} SA. Best viewed in color.
  }
  \label{fig:grad_cam}
\end{figure*}

\noindent \textbf{Visualization of attention maps.}
To intuitively understand how the SA module achieves semantics adaption from pre-training to the application domain, we visualize the attention maps in Fig.~\ref{fig:grad_cam}.
We observe that \textbf{(1)} Since Market-1501 is relatively easy with greater inter-class differences, the model can leverage learned general human semantics to focus on the human body. As shown in the left of (b), the model can distinguish pedestrians based on their different body semantics. When adopting SA, as shown in the left of (c), the model concentrates more on discriminative body regions and does not rely on the face, which improves its robustness since the face is usually not considered the dominant information in Re-ID. Paying excessive attention to it can be misleading due to low resolution, view changes, or occlusions.
\textbf{(2)} MSMT17 is much more challenging with greater variances. Without semantics adaption, the model is susceptible to background or fragile person clues, \textit{e.g.}, small logos, or corners of clothes, as shown in the right of (b). However, the SA module helps to concentrate on discriminative regions of the human body, effectively promoting Re-ID. 
Overall, our proposed SA module can handle different domains with great discrepancy, showing great ability and prospect to evolve the model for new deployment and wide application.

\subsection{Limitations and Future Work}
Our proposed LReID paradigm leverages the pre-trained models, which capture robust and general human semantic knowledge after being trained on diverse and challenging pedestrian images at a large scale. Although we show that current pre-trained models have been capable of supplying useful semantics for lifelong evolution, collecting more diverse and less biased data for better pre-training is expected to expand their capacity.

\section{Conclusion}
In this paper, we propose the Distribution Aligned Semantics Adaption (DASA) framework that is free of exemplars and old models to effectively take advantage of the pre-trained model for LReID.
We assume the Re-ID model pre-trained with large-scale pedestrian data can acquire robust and general human semantics as shared knowledge for lifelong application. To effectively adapt the pre-trained model to each application domain, we propose to perform distribution alignment by tuning BN and propose the lightweight Semantics Adaption (SA) module to aggregate learned semantics for better pedestrian representations. Extensive experiments show the superiority of the proposed framework for LReID in storage consumption and performance. It is expected to inspire more research to make full use of pre-trained models for effective and efficient LReID, making the landing application of Re-ID possible.

\bmhead{Acknowledgements}
This work was supported in part by the NSFC Project (62176061), in part by the Shanghai Municipal Commission of Economy and Informatization (2021-GZL-RGZN-01033), in part by the Shanghai Research and Innovation Functional Program (17DZ2260900), and in part by the Shanghai Technology Development and Entrepreneurship Platform for Neuromorphic and AI SoC.

\bibliography{sn-bibliography}

\end{document}